\documentclass[10pt,twocolumn,letterpaper]{article}

\usepackage[accsupp]{axessibility}  % Improves PDF readability for those with disabilities.

\usepackage{iccv}
\usepackage{times}
\usepackage{epsfig}
\usepackage{graphicx}
\usepackage{amsmath}
\usepackage{amssymb}

\usepackage{algorithm}
\usepackage{algorithmic}
\usepackage{amsmath}
\usepackage{bbm}
\usepackage{mathtools}
\usepackage{multicol}
\usepackage{multirow}
\usepackage{tikz}
\usepackage{booktabs}
\usepackage{amsthm}

\usepackage{authblk}

\usepackage{subcaption}

\usepackage{xcolor}

\usepackage[breaklinks=true,bookmarks=false]{hyperref}

\iccvfinalcopy % *** Uncomment this line for the final submission

\def\iccvPaperID{10216} % *** Enter the ICCV Paper ID here

\ificcvfinal\fi

\newcommand{\bftab}{\fontseries{b}\selectfont}

\newtheorem{theo}{Theorem}

 \def \x{\mathbf x}
 \def \W{\mathbf W}
 \def \w{\mathbf w}
 \def \b{\mathbf b}
 
\newcommand{\norm}[1]{\left\lVert#1\right\rVert }

\newcommand{\DS}{\emph{DenseShift} }

\newcommand{\Sthree}{S$^3$~}

\def\wshift{w_\mathrm{shift}}

\def\wsign{w_\mathrm{sign}}

\def\cm{\multicolumn{1}{c}{\checkmark}}
\def\d{\multicolumn{1}{c}{$-$}}
\tikzstyle{arrow} = [thick,->,>=stealth]
\def\checkmark{\tikz\fill[scale=0.4](0,.35) -- (.25,0) -- (1,.7) -- (.25,.15) -- cycle;}

\def\Val{{\sf{fl}}}

\def\ADDint{{\sf{Add}_{\sf{uint}}}}

\def \real{{\rm I\!R}}

\begin{document}

%%%%%%%%% TITLE
% \title{\DS: Towards Accurate and Transferable Low-Bit Shift network}
\title{\DS: Towards Accurate and Efficient \\ Low-Bit { Power-of-Two Quantization} }

\author[1]{\bftab Xinlin Li}
\author[2]{\bftab Bang Liu}
\author[1]{\bftab Rui Heng Yang}
\author[1]{\bftab Vanessa Courville}
\author[1]{\bftab Chao Xing}
\author[1]{\bftab Vahid Partovi Nia}
\affil[1]{Noah’s Ark Lab, Huawei Technologies.}
% \affil[ ]
% {\textit {\{xinlin.li1, rui.heng.yang1, vanessa.courville, xingchao.ml, vahid.partovinia\}@huawei.com}
% }
\affil[2]{University of Montreal \& Mila - Quebec AI Institute}
\affil[1]
{
\textit {\{
\href{mailto:xinlin.li1@huawei.com}{xinlin.li1}, 
\href{mailto:rui.heng.yang1@huawei.com}{rui.heng.yang1}, 
\href{mailto:vanessa.courville@huawei.com}{vanessa.courville}, 
\href{mailto:xingchao.ml@huawei.com}{xingchao.ml}, \href{mailto:vahid.partovinia@huawei.com}{vahid.partovinia}
\}@huawei.com}
}
\affil[2]
{
\textit{\href{mailto:bang.liu@umontreal.ca}{bang.liu@umontreal.ca}
}
}

% \author{Xinlin Li\\
% Huawei Noah’s Ark Lab\\
% \and
% Bang Liu\\
% {\tt\small secondauthor@i2.org} \\ 
% \and Rui Heng Yang \\
% Huawei Noah’s Ark Lab \\
% \and  Vanessa Courville \\
% Huawei Noah’s Ark Lab \\
% \and  Chao Xing \\
% Noah’s Ark Lab, Huawei Technologies. \\
% \and  Vahid Partovi Nia \\
% Huawei Noah’s Ark Lab \\

\maketitle
\ificcvfinal\thispagestyle{empty}\fi

%%%%%%%%% ABSTRACT
\begin{abstract}

%Deploying deep neural networks on low-resource edge devices is challenging due to their ever-increasing resource requirements. 
%Recent investigations propose multiplication-free neural networks to reduce computation and memory consumption. Shift neural networks is one of the most effective tools towards these reductions. 
%However, existing low-bit Shift networks are not as accurate as their full precision counterparts and cannot efficiently transfer to a wide range of tasks due to their inherent design flaws.
%We propose \DS network that exploits the following novel designs.
%First, we demonstrate that the zero-weight values in low-bit Shift networks are neither useful to the model capacity nor simplify the model inference. Therefore, we propose to use a zero-free shifting mechanism to simplify inference while increasing the model capacity.
%Second, we design a new metric to measure the weight freezing issue in training low-bit Shift networks, and propose a sign-scale decomposition to improve the training efficiency.
%Third, we propose the low-variance random initialization strategy to improve the model's performance in transfer learning scenarios.
%We run extensive experiments on various computer vision and speech tasks.
%The experimental results show that DenseShift network significantly outperforms existing low-bit multiplication-free networks and can achieve competitive performance to the full-precision counterpart. 
%It also exhibits strong transfer learning performance with no drop in accuracy.

Efficiently deploying deep neural networks on low-resource edge devices is challenging due to their ever-increasing resource requirements. To address this issue, researchers have proposed multiplication-free neural networks, such as Power-of-Two quantization, or also known as Shift networks, which aim to reduce memory usage and simplify computation. However, existing low-bit Shift networks are not as accurate as their full-precision counterparts, typically suffering from limited weight range encoding schemes and quantization loss. In this paper, we propose the \DS network, which significantly improves the accuracy of Shift networks, achieving competitive performance to full-precision networks for vision and speech applications.  In addition, we introduce a method to deploy an efficient \DS network using non-quantized floating-point activations, while obtaining $1.6\times$ speed-up over existing methods. 
To achieve this, we demonstrate that zero-weight values in low-bit Shift networks do not contribute to model capacity and negatively impact inference computation. To address this issue, we propose a zero-free shifting mechanism that simplifies inference and increases model capacity.
We further propose a sign-scale decomposition design to enhance training efficiency and a low-variance random initialization strategy to improve the model's transfer learning performance.
Our extensive experiments on various computer vision and speech tasks demonstrate that \DS outperforms existing low-bit multiplication-free networks and achieves competitive performance compared to full-precision networks. Furthermore, our proposed approach exhibits strong transfer learning performance without a drop in accuracy.
{ Our code was released on
\href{https://github.com/xinlinli1990/noah-research/tree/master/S3-Training}{GitHub}.}

\end{abstract}

% \begin{abstract}
    
% Efficiently deploying deep neural networks on low-resource edge devices is challenging due to their ever-increasing resource requirements. To address this issue, researchers have proposed multiplication-free neural networks, such as low-bit Shift networks. The Shift network is useful to accelerate inference, but often requires dedicated hardware. \DS improves Shift network inference in two directions i) allows it to be implemented with floating-point activation; ii) speeds up inference $1.6\times$.  We also propose a new initialization method to improve \DS training and beat state-of-the-art shift networks  in terms of accuracy.  
% We demonstrate that zero-weight values in low-bit Shift networks do not contribute to model capacity and negatively impact inference computation. Our extensive experiments on various computer vision and speech tasks demonstrate that \DS outperforms existing low-bit multiplication-free networks, and achieves competitive performance compared to full-precision networks. Furthermore, our proposed approach exhibits strong transfer learning performance with no drop in accuracy.

% \end{abstract}

%%%%%%%%% BODY TEXT

\section{Introduction}

\begin{figure}[t]
    \centering
    \includegraphics[width=0.85\linewidth]{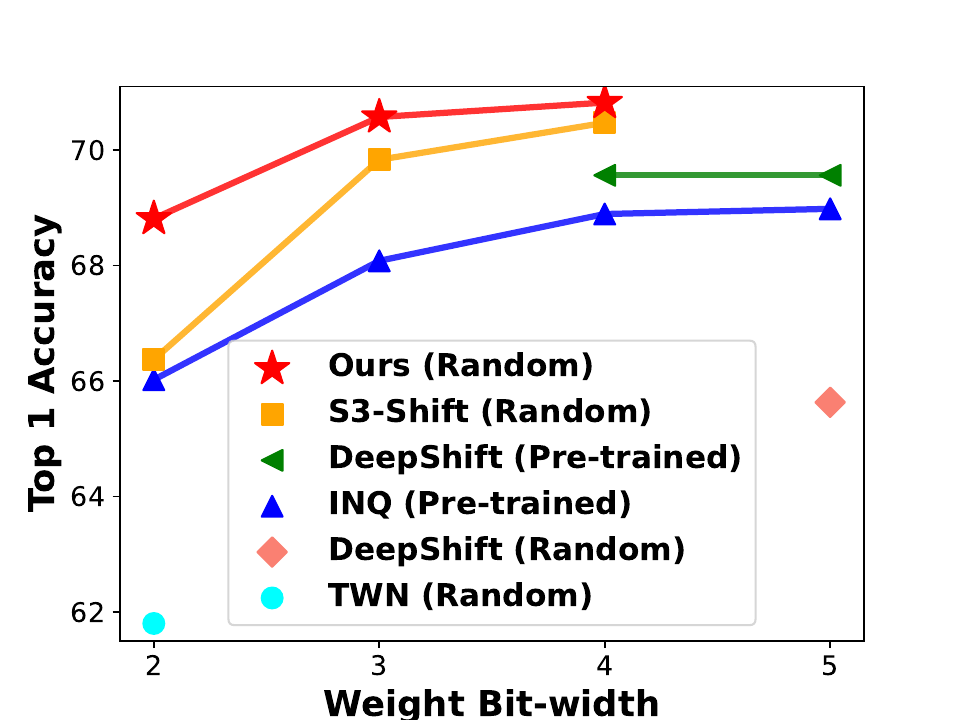}
    \caption{Benchmark low-bit DenseShift networks over SOTA low-bit Shift networks on ImageNet using the ResNet-18 model architecture. 
}
 % \vspace{-4mm}
	\label{fig:ResNet-18-ImageNet-Top1}
\end{figure}

% DenseShift network over Shift network
% 1)  
% 2) 

Deep neural networks have demonstrated superior performance in diverse applications such as image classification, object detection, and image segmentation \cite{he2016identity,liu2016ssd,chen2017rethinking}, and speech \cite{lugosch2019speech}. However,  despite the high accuracy of multiplication-based deep neural networks, their computing resource requirement makes their deployment challenging, especially on low-resource devices. Recent research has explored multiplication-free neural networks that reduce memory footprint and overall energy consumption to address this issue.

%However, this performance comes at the cost of increased model size, which leads to a higher computational complexity due to the increased amount of floating-point multiplication in training and inference time, yielding a higher memory footprint and increasing energy consumption. To reduce computational complexity, researchers are exploring multiplication-free neural networks. 
% \red{Please check if following paragraph helps}
%\blue{Despite the excellent performance gained by the multiplication-based deep neural networks, the complexity of the models make deploying such models on devices, especially edge devices challenging. Therefore, researchers are exploring multiplication-free neural networks to reduce the memory footprint and decrease the energy consumption of multiplication-based models. }

Existing works on multiplication-free neural networks include binary \cite{courbariaux2015binaryconnect}, and ternary quantization \cite{li2016ternary}. They respectively constrain their weights in the range of $\{\pm1\}$ and $\{0\} \cup \{\pm1\}$, in order to replace multiplication computations with less expensive operations such as a sign flip operator. These low-bit quantization techniques make it possible to deploy deep learning models on resource-constrained edge devices.
Moreover, \cite{wang2021addernet} trades the multiplication operation with the addition operation, and  \cite{elhoushi2019deepshift,li2021s,fbsaving} use the bit-shift operator to build power-of-two (PoT) quantized networks, known as Shift networks.
Shift networks built upon a ternary base have a weight space of $\{0\} \cup \{\pm 2^p\}$. This means that multiplication operations can be replaced with bit-wise shift operations, which have highly efficient hardware implementations. 
In fact, \cite{fbsaving} showed that with 4-bit weights and 8-bit activations, the shift-based MAC unit designed for Shift networks outperformed its counterpart for traditional uniform quantization by $2.4\times$ energy saving and 20\% chip area saving using Samsung 5nm.

A recent study \cite{li2021s} proposes a weight reparameterization scheme $S^3$ for Shift network training, which significantly improves the accuracy of the ImageNet classification task under sub-4bits weight and does not require full-precision pre-training. % It shows that the poor performance of previous Shift networks are caused by the design of the ternary quantizer, which prevents the discrete weight value across zero, leading to their weight sign cannot easily change during training. This phenomenon is called weight freezing.
However, $S^3$ has the following shortcomings:
% i) $S^3$ exploits a low weight bit-width, which causes performance drop under low-bit conditions compared to full-precision networks;  
% % \red{Shortcomings?}
i)  Existing Shift networks, including $S^3$, only support quantized activations during inference, limiting their performance gains and usefulness in various scenarios;
% % ii) $S^3$ fails to explain the weight freezing problem for non-ternary quantizers, and the handling of zero weights increases the computational complexity during inference;
ii) $S^3$ is only benchmarked on image classification and exhibits significant performance degradation under 2-bit weight; iii) Transfer learning tasks are unexplored.

In this study, we identify and address design limitations in current low-bit Shift networks through a detailed analysis, resulting in the proposal of \textit{DenseShift} network. Our novel designs significantly enhance model capacity, inference efficiency, and transferability. The contributions of this study are outlined below.

First, our analysis reveals that zero weights in low-bit Shift networks reduce model capacity under limited bit widths. To address this issue, we propose a zero-free shifting mechanism that removes zero values from the weight space. This design enhances model representation capacity and improves performance under low-bit conditions, surpassing existing low-bit Shift networks.

Second, we introduce a novel inference approach for \DS networks that supports both floating-point and quantized activations. Our approach accelerates the dot-product computation by $1.6\times$ on ARMv8 CPU under FP16. Notably, \DS is the first Shift network that enables inference with non-quantized floating-point activations and the first to demonstrate performance improvement without relying on dedicated hardware such as ASIC or FPGA.

Third, we propose an efficient training algorithm for \DS networks adapted from the weight re-parameterization techniques \cite{li2021s}. Our sign-scale decomposition method breaks down the discrete weights into a binary sign term and a power-of-two scale term, and recursively re-parameterizes the exponent of the scale term as a combination of binary variables. This enables us to train low-bit \DS networks from a random initialization, achieving performance that is comparable to full-precision networks.

Fourth, while prior research works suffer from severe performance degradation when transferred to a new task, we propose a low-variance random initialization strategy to improve the model’s performance in transfer learning scenarios. 
We demonstrate that the weight values tend to gather towards the original point of the re-parameterization space during the initial stage of training, and as a result, a greater gradient signal is needed to push them to pass the threshold when the weights are randomly initialized with a large variance. 
By reducing the variance of weight initialization, the DenseShift network can be easily adapted to different tasks while maintaining competitive performance. 

We conducted extensive experiments to evaluate the performance of our DenseShift network compared to various baselines on a diverse set of tasks across different fields. The results show that our proposed DenseShift network outperforms the state-of-the-art Shift network on the ImageNet classification task and achieves comparable performance to full-precision networks while having higher inference computational efficiency. 
As summarized in Fig. \ref{fig:ResNet-18-ImageNet-Top1}, DenseShift network performs significantly better in low-bit settings, especially under 2-bit condition. Specifically, our { 2-bit and} 3-bit quantized ResNet-18 on the classification task achieve  $68.90\%$ and $70.57\%$ Top-1 accuracy respectively.
Moreover, we demonstrate that our low-bit DenseShift networks can achieve full-precision performance in transfer learning scenarios across different domains, including computer vision and speech tasks.  This study is the first to demonstrate this capability, to the best of our knowledge.

\section{Related Works}
% \textbf{Multiplication Free Network}
Different approaches have been suggested to replace the expensive multiplication operation to mitigate the computational complexity of neural networks. 
Low-bit neural networks with binary weights \cite{courbariaux2015binaryconnect,rastegari2016xnor} or ternary weights \cite{li2016ternary} are examples of multiplication-free networks. While computationally inexpensive, their major flaw lies in the accuracy gap compared to their full-precision counterparts, as they suffer from under-fitting on large datasets. 
There are also works that utilize computationally cheaper operations, such as addition operations \cite{chen2020addernet,wang2021addernet,xu2020kernel}, square operations \cite{prazeres2021euclidnets}, or bit-shift operations \cite{zhou2017incremental,gudovskiy2017shiftcnn,elhoushi2019deepshift,miyashita2016convolutional}. Compared to using binary or ternary weights, these methods achieve a low accuracy drop on large datasets but require higher weight representation bit-width as a trade-off. 
Some other works try to improve the performance of multiplication-free neural networks by using both addition and bit-shift operations \cite{you2020shiftaddnet}, a sum of binary bases \cite{lin2017towards,zhang2018lq}, or sum of shift kernels \cite{li2019additive}, however, they remain computational costly as more operations are used per kernel.

In order to improve the accuracy of Shit networks under low-bit, \cite{zhou2017incremental} propose to fine-tune the pre-trained full-precision weights with a power-of-two quantizer in a group-by-group manner. \cite{elhoushi2019deepshift} proposed a power-of-two quantizer design which allows training Shift networks from scratch. However, initialization with a pre-trained full-precision checkpoint is still critical for achieving high accuracy under low-bit.
\cite{li2021s} proposes a weight reparameterization technique $S^3$ for training low-bit Shift networks. It points out the design flaw of the weight quantizer for low-bit Shift networks and proposes to decompose a discrete parameter in a sign-sparse-shift 3-fold manner to improve ImageNet classification accuracy under sub-4bits conditions significantly and no longer rely on full-precision pre-training.
% \cite{li2021s} proposes a weight reparameterization technique $S^3$ for training low-bit Shift networks. It points out the design flaw of the weight quantizer for low-bit Shift networks and proposes to decompose a discrete parameter in a sign-sparse-shift 3-fold manner to improve training efficiency.
% Compared to \cite{li2021s}, the zero-free weight space in our proposed \DS network further improves the model representation capacity and inference efficiency. 
% And the low-variance random initialization strategy equips low-bit Shift networks with strong transferability, which is what $S^3$ lacks.

\begin{figure}[t]
    \centering
    \label{fig:Inference-shift-float}
    \includegraphics[trim=9 390 574 0,clip,width=\linewidth]{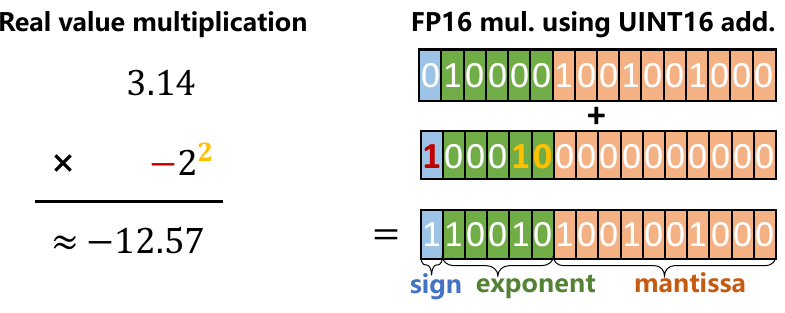}
\caption{The multiplication between a float number and a positive or negative power-of-two integer can be implemented by an integer addition instruction, which allows \DS networks inference on most existing hardware efficiently. The FP16 dot-product computation achieved $1.6 \times$ speed up on ARM A57 CPU using this technique as discussed in Sec. \ref{sec:non-quantized-inference}.}
\label{fig:DenseShift_Float}
\end{figure}

\section{\DS Network}

The following section introduces the proposed \DS network, highlighting the benefits for inference deployment, and providing a detailed analysis of weight encoding space, training mechanisms and weight initialization strategies employed.

\subsection{\textit{DenseShift} with Zero-Free Shifting}
% We formally introduce \DS network, a transfer-learnable zero-free Shift network designed for low-bit conditions. 

% In typical Shift networks, a weight space with $n$-bits can encode up to $2^n$ discrete weight values. These values are usually centered around zero, requiring a weight value of zero to be included in this encoding scheme. However, since the zero value has no corresponding negative value, the encoding space utilization rate is reduced to $\frac{2^n-1}{2^n}$. Under the high-bit conditions, this waste ratio is insignificant and is often ignored. However, under low-bit, especially when $n\leq 4$, this additional encoding space plays a crucial role. 
Typical Shift networks use a weight space with $n$-bits to encode weight values, allowing up to $2^n$ discrete values. However, since the values are usually centred around zero, the utilization rate of the encoding space is reduced when zero is included. This becomes significant under low-bit conditions, particularly when $n\leq4$.

%For example, the encoding rate is reduced to only $75\%$ in a 2-bit weight space.
% A weight space of $n$-bits can encode up to $2^n$ discrete weight values. Since the zero value has no corresponding negative value, the encoding space utilization rate is reduced to $\frac{2^n-1}{2^n}$ in classical Shift networks. In high-bit, this waste ratio is insignificant and is often ignored, but in low-bit conditions when $n\leq 4$ even one bit saving plays a crucial role.
% , e.g. for $n=2$ the encoding loss is $75\%$. \DS uses this flaw towards encoding more values in fewer bits to fully-utilize the encoding space. Consequently, \DS networks can significantly outperform Shift networks under limited model capacity such as in low-bit weight conditions.
% \DS , on the other hand, fully utilizes the entire encoding space by dropping the zero value. 
Taking $n=2$ bits as an example, the weight space allows for $4$ discrete weight values to be encoded. In a typical Shift network, these would include $w=\{-1,0,+1\}$, ignoring the potential for adding a fourth weight value. In \DS however, as there is no zero-value, we can now encode weight values of  $w=\{-2, -1, +1, +2\}$ with the same number of bits. This increases the overall range of weight values supported, allowing \DS to significantly outperform existing Shift networks, especially under low-bit weight conditions as summarized in Fig. \ref{fig:ResNet-18-ImageNet-Top1}.

% { Although the zero-free shifting design makes the model incompatible with the non-structural pruning technique \cite{han2015deep}}, we argue that it is impractical to prune a low-bit DenseShift network with the number of bits $n\leq 4$. That is because the sparse matrix format is inefficient in the low-bit situation as it stores non-zero values and indexes them in pairs. 
% For instance, a typical CNN kernel size is $3\times 3\times 256 \times 256>2^{16}$. Each sub-4bits weight value may require a uint32 index, making memory {usage} inefficient.

\subsubsection{Inference with floating-point activation}
\label{sec:non-quantized-inference}
The Zero-Free Shifting design also brings additional benefits to inference. To the best of our knowledge, all existing Shift networks rely on using quantized activations in order to effectively replace the multiplication between the power-of-two weights and activations with bit-shift instructions. However, there are some challenges with quantized activations when applied in practice. For instance, in LSTM and Transformer models, many operators, such as softmax or addition, are unable to compute directly with the quantized activations. Instead, it must first dequantize in order to compute, then re-quantize the results, which leads to extra inference latency.
% and is user-unfriendly to the regular AI developer without quantization knowledge. 
This is also seen with large language models (LLM) as shown in \cite{dettmers2022llm} where 8-bit quantization could not maintain full-precision performance on LLMs with models that exceeded 6.7 billion parameters because the fixed-point quantization can not handle activations with a large dynamic range well, leading to a significant accuracy degradation of the LLMs. 
% It's shown that as the number of model parameters increases, outlier features emerge during inference, thus increasing the dynamic range of the activations. The fixed-point quantization can not handle activations with a large dynamic range well, leading to a significant accuracy degradation of the LLMs. 
These issues above impair the performance gains or usability of the existing power-of-two quantization methods.

In this work, we propose a method to calculate the multiplication between a floating-point number and positive or negative power-of-two numbers using integer addition instructions. Our approach allows \DS networks to perform inference directly on non-quantized floating point activations, thus avoiding the above issues. 

In the following section, we describe how to achieve equivalent multiplication between floating-point numbers and power-of-two numbers using lower-bit integer addition. A floating-point number is obtained by Eq. \ref{eq:float_val}.

\begin{equation} 
    \label{eq:float_val}
    % {\textbf{Val}}_{float} 
    % \Val(x) = (-1)^{\textbf{s}} \times 2^{\textbf{e} + \textbf{e}_{\textbf{bias}}} \times (1 + \frac{\textbf{m}}{2^{\textbf{m}_{\textbf{bits}}} - 1}),  
    \Val(x) = (-1)^{x_{\textbf{s}}} \times 2^{x_{\textbf{e}} + \textbf{e}_{\textbf{bias}}} \times (1 + \frac{x_{\textbf{m}}}{2^{\textbf{m}_{\textbf{bits}}} - 1}),  
\end{equation}

Where $\Val(\cdot)$ is the float representation, $x_{\textbf{s}}$ is the sign bit value, $x_{\textbf{e}}$ is the unsigned integer value represented by the exponent bits, $x_{\textbf{m}}$ is the unsigned integer value represented by the mantissa bits, and $\textbf{m}_{\textbf{bits}}$ is the mantissa bit-width and $\textbf{e}_{\textbf{bias}}$ the constant exponent bias value defined in the floating-point standard. For the 32-bit float format defined in the IEEE 754 standard, $\textbf{m}_{\textbf{bits}}=23$ and $\textbf{e}_{\textbf{bias}}=-127$.

From the floating-point representation, it is clear that the multiplication of a floating-point value with a power-of-two number is equivalent to adding a corresponding integer to the exponent bit of the floating-point number. The negation of a floating-point number is equivalent to performing a bit-flip on the sign bit, which can be achieved by adding one to the sign bit. Therefore, the multiplication of a float number with a positive or a negative power-of-two number can be performed by one single lower-bit integer addition operation on its sign and exponent bits as described in Fig. \ref{fig:DenseShift_Float}.
As an example, the multiplication between a 32-bit float number and a positive or negative power-of-two integer can be implemented with a 9-bit integer adder on its sign bit and exponent bits. 
Related works \cite{horowitz20141,sentieys2021approximate} show that replacing floating-point multiplication with fixed point addition can save $37\times$ energy cost and $56\times$ chip area cost at 32-bit, and using an 8-bit integer adder can further reduce $3.3\times$ energy cost and $3.8\times$ chip area. This highlights the potential of \DS networks to reduce power consumption and chip area for AI chips.
% According to related works \cite{horowitz20141,sentieys2021approximate}, replacing floating-point multiplication with fixed point addition saves 37X energy cost and 56X chip area cost at 32-bit. Moreover, the 8-bit integer adder can further reduce 3.3X energy cost and 3.8X chip area compared to the 32-bit integer adder. This indicates that the \DS network has great potential to reduce the AI chip power consumption and the chip area.

Furthermore, our proposed \DS inference approach is compatible with the existing hardware. Thanks to its Zero-Free Shifting design, the multiplication instruction in the \DS inference can be replaced by one single integer addition instruction which requires fewer execution cycles in general as described in Eq. \ref{eq:float_mul} and Fig. \ref{fig:DenseShift_Float}.

% \begin{equation} 
%     \label{eq:float_mul}
%     \textbf{MULT}_{float}({\textbf{Val}}_{float}, \pm2^{p}) = \textbf{ADD}_{uint}(\textbf{Val}_{float}, \pm2^{p+\textbf{e}_{\textbf{bias}}})
% \end{equation}
\begin{equation} 
    \label{eq:float_mul}
    \Val(2^p x) = \ADDint(\Val(x) , \Val(2^{p+\textbf{e}_{\textbf{bias}}}))
\end{equation}
% \begin{equation} 
%     \label{eq:float_mul}
%     \textbf{MULT}_{float}({\textbf{Val}}_{float}, \pm2^{p}) = \textbf{ADD}_{uint}(\textbf{Val}}_{float}, \pm2^{p+n})
% \end{equation}
% where , and $\n$ the exponent bias of the float representation of $x$,  defined in the floating-point standard and the $\Uint$ operator attaches the exponent and matinssa blocks to treat this combined bits as a single UINT. 

% \begin{equation} 
%     \label{eq:float_mul}
%     \textbf{MULT}({\textbf{Val}}_{float}, \pm2^{p})_{float} = \textbf{ADD}(\textbf{Val}_{{float}}, \pm2^{p+\textbf{e}_{\textbf{bias}}})_{{int}}
% \end{equation}

% \begin{equation} 
%     \label{eq:float_mul}
%     ({\textbf{Val}}_{float} \times \pm2^{p})_{\textbf{float}} = (\textbf{Val}_{float} + \pm2^{p+\textbf{e}_{\textbf{bias}}})_{\textbf{int}}
% \end{equation}
Where $\Val(\cdot)$ is the float representation, $\textbf{e}_{\textbf{bias}}$ the constant exponent bias value defined in the floating-point standard. $\ADDint(\cdot,\cdot)$ is the integer addition.

Unlike the UINT8 and UINT16 formats, widely supported by existing hardware, the FP8 and FP16 formats have relatively limited hardware support. This is because floating-point numbers have the disadvantage of significant calculation error under lower bits, which can not satisfy the needs of many computing tasks. As a result, most existing hardware do not support low-precision floating-point arithmetic. They are reserved for data storage or only supported by limited operators. 
For instance, ARMv8 and X86 AVX2 instruction sets do not support FP16 arithmetic. When an operation is required, it is necessary to convert the FP16 to FP32 in registers before the operation and convert it back afterward. The same strategy is used by NVIDIA GPUs while processing FP8 \cite{cuda}.
On this hardware, we can use the corresponding unsigned int addition to replace the floating-point multiplication instruction, which has the potential to achieve inference speed up. 
{ It's important to note that our proposed approach is not applicable to zero-multiplication, as it necessitates additional operations for implementation, making it less efficient on general hardware. This underscores the advantages of our zero-free shifting design.}

% since addition instruction requires fewer cycles for execution in general.

% Our proposed inference approach improves the infernce energy efficiency at the same time. According to related works \cite{horowitz20141, sentieys2021approximate}, 16 bit integer addition can achieve 22X energy cost saving and 24.5X chip area

% TO DO: ARM experiments$

As a proof-of-concept, we implement \DS using this technique and compare it to floating-point dot-product using a vectorized software implementation.
% {\color{blue}add an equation to illustrate similar to equation 1?}
Both weights and activations are provided to the compute kernel as FP16.
% , which contains 1 bit for the sign, 5 bits for the exponent and 10 bits for the mantissa
The \DS kernel performs bit-wise manipulation and adds the sign and exponents of the weight and activation values together (see Fig. \ref{fig:DenseShift_Float}), effectively replacing the floating-point multiplication with a simple integer addition. The result of this integer addition is then cast back to FP32 for accumulation. This implementation is compared to the dot-product baseline adapted from the FP16 GEMM kernel of the open-source inference library NCNN \cite{Ni_ncnn_2017} for the ARMv8 hardware.
% {Mentioned NCNN baseline instead to strengthen our claim ?}
% {a half-precision dot-product which receives both the half-float weights and half-float activations and performs a multiply-accumulate operation. Note that due to limitations of the hardware, the multiply-accumulate is performed in 32-bit float as well. }
All experiments were run on ARM A57 CPU using NEON SIMD architecture and count the average time consumption.  We run experiments for 4096 data points. The results averaged over 1000 runs show that the latency for the floating-point dot-product and our proposed technique are $5.98 \mu s$ and $3.76 \mu s$, respectively. In other words, \DS kernel using the proposed floating-point technique obtains \textbf{1.6× speed-up}. % compared to the baseline floating-point dot-product implementation. 

Our \DS implementation can be further optimized to reduce overall memory consumption requirements; since the weight values are constrained to power-of-two numbers, their mantissa will always be zero and thus are not needed in the compute kernel. Instead, only the weight's sign and exponent are sent to the compute kernel, requiring only 7 bits, and can be represented with a single unsigned 8-bit integer. %, reducing overall memory consumption requirements.
In addition, our proposed approach has the potential to be extended to other neural network layers beyond matrix multiplication for efficient computation.

% \cite{you2020shiftaddnet}, compared to 32-bit floating-point multiplication, its integer addition counterpart can achieved 31X energy saving on 45nm ASIC and 196X energy saving on FPGA.

% Unlike the existing Shift networks, the \DS weights only contain positive and negative power-of-two numbers and exclude zeroes. This design allows the \DS network to 
% directly compute multiplication on non-quantized floating-point activation using integer addition instruction by leveraging the numerical format definition of floating-point numbers. 

\subsubsection{Inference with quantized activation}

\begin{figure}[t]
\centering
\subcaptionbox{Shift MAC kernel
    \label{fig:Inference-shift-ip}}
    {\includegraphics[trim=20 270 720 0,clip,width=.54\linewidth]{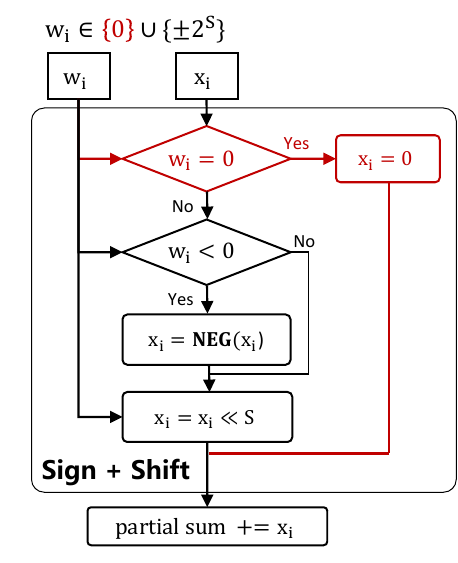}}
    \hspace*{0.1cm}
\subcaptionbox{DenseShift MAC kernel
    \label{fig:Inference-denseshift-ip}}
    {\includegraphics[trim=20 290 780 0,clip,width=.41\linewidth]{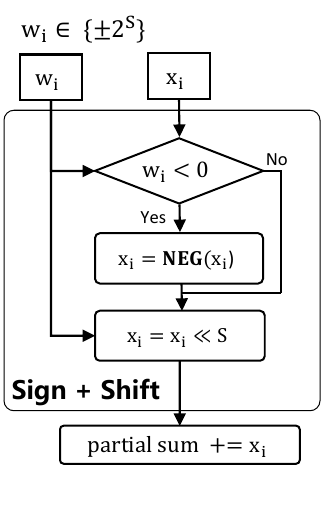}}
  \vspace{-3mm}  
\caption{Compare the Multiply-Accumulate (MAC) operations in Shift and DenseShift for quantized activation. The DenseShift MAC is more efficient as its $w_{i}$ excludes zero.}
 \vspace{-5mm}
\label{fig:Inference}
\end{figure}

With a zero-free weight space design, \DS simplifies the inference computation on fixed-point quantized activations as well. {Figure \ref{fig:Inference} compares the Multiply-Accumulate (MAC) operation between \DS and Shift networks.
% We can observe from Fig. \ref{fig:Inference-shift-ip} that the fixed-point dot product operation in existing Shift networks receives two fixed-point input vectors: the activations $x_i$, and the weights $w_i$ that are encoded by its shift value $S$.}
% The shift operation is performed by first sign-flipping $x_i$, shifting the result by its corresponding shift value $S_i$, and accumulating over all data points. 
% However, special handling is required when $w_i=0$ to bypass these operations and simply pass a value of $x_i=0 $ to the accumulator instead. 
The Shift network MAC kernel requires special handling when $w_i=0$, as shown in Fig. \ref{fig:Inference-shift-ip}, which bypasses sign-flip and bit-shift operations and pass a value of $x_i=0 $ to the accumulator instead.
Since \DS guarantees that $w_i$ will never be zero, this branch is no longer required, as shown in Fig. \ref{fig:Inference-denseshift-ip}. Therefore, the inference computation in \DS networks is simpler and more efficient than in existing Shift networks.
%\blue{We also show the computational efficiency of \DS network by comparing with existing Shift networks through a vectorized software implementation. }
% To demonstrate this advantage in computational efficiency, we compare \DS  to existing Shift networks using a vectorized software implementation. We implemented the \DS kernel and \textit{Shift} kernel for dot-product computation on ARM A57 CPU using NEON SIMD architecture and count the average time consumption. Due to the limitations of ARM neon intrinsic data types, we consider 8-bit weights as a proof-of-concept. We run experiments for 4096 data points. The results averaged over 1000 runs show that the latency for the Shift network and our DenseShift network are $2.65us$ and $1.79us$, respectively. In other words, the \DS kernel obtains 1.48X speed-up compared to a \textit{Shift} kernel implementation.  
A NEON SIMD dot-product kernel was developed on an ARM A57 CPU to demonstrate \DS's computational efficiency over existing Shift networks. The experiments were performed with INT8 for 4096 data points, and the results showed that \DS kernel had a $1.48\times$ speed-up compared to a \textit{Shift} kernel implementation, with the latency of $1.79 \mu s$ and $2.65 \mu s$, respectively.

% \subsubsection{Theory}

% Aside from experimental demonstration, we also theoretically show that removing zeros from the weight space doesn't affect the representation power of DenseShift models.
% The theorems are shown as follows, and the detailed proofs of them are described in the supplementary material:
Aside from experimental demonstration, we also theoretically show that removing zeros from the weight space doesn't affect the representation power of DenseShift models, see the Supplementary Material. 
Thoerem~1  confirms that there is a DenseShift network that can reach to the same accuracy as any Shift network if properly trained.  Theorem~2 shows there is a \DS network that can attain the same capacity compared with a full-precision network. 

\subsection{Sign-Scale Decomposition for Efficient Training}

\begin{figure}[t]
	\centering
	\includegraphics[trim=5 362 780 0,clip,width=0.49\linewidth]{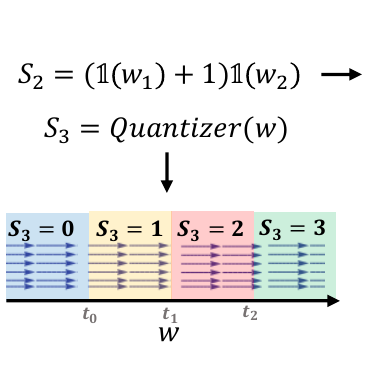}
	\includegraphics[trim=10 370 780 0,clip,width=0.49\linewidth]{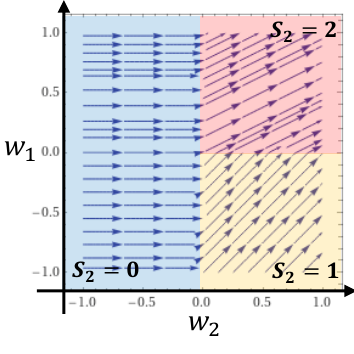}
	\includegraphics[trim=0 330 770 10,clip,width=0.45\linewidth]{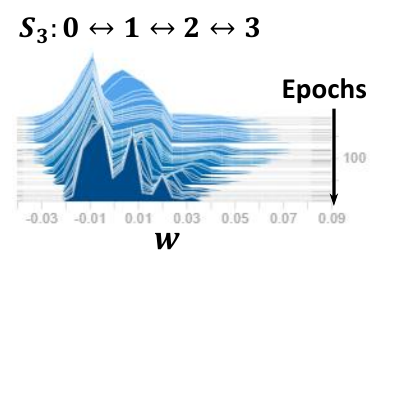}
	\includegraphics[trim=30 0 90 40,clip,width=0.5\linewidth]{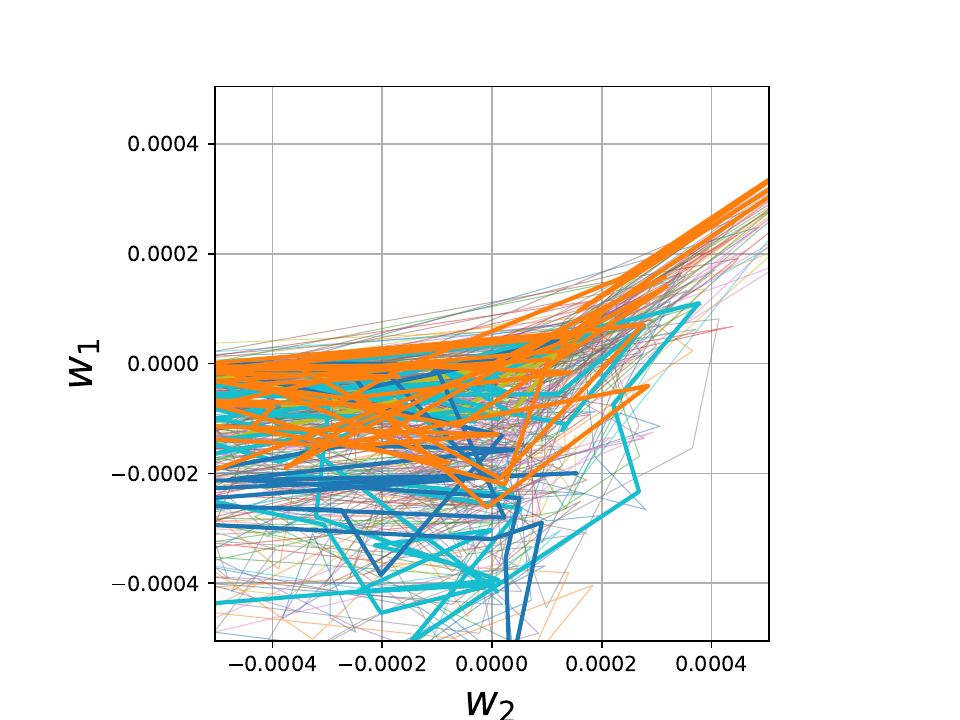}
	\caption{
	Top Left: The optimization space of the shifting parameter $S_3$ defined by the sign-shift quantizer adapted from \cite{elhoushi2019deepshift}. The arrow direction represents the moving direction of the continuous weight $w$ in the optimization space when the gradient update of the shifting parameter $S_T$ is positive. 
	Bottom Left: The actual continuous weight histogram variation of $S_3$ defined by the 3-bit sign-shift quantizer during training. 
	Top Right: The phase plane of the re-parameterization space of $S_2$ defined by a recursive product of binary variables ($\frac{\partial w_2}{\partial t}=\mathbbm{1}(w_{1})+1, \frac{\partial w_1}{\partial t}=\mathbbm{1}(w_{2})$). 
	Bottom Right: Some sampled actual weight traces of $S_2$ on the optimization space defined by a recursive product of binary variables during training. The weight trajectories show that the $S_2$ values can quickly move among multiple discrete values during training and are no longer limited to the two adjacent values of the quantizer's thresholds.
	}
	
	 % \vspace{-4mm}
	\label{fig:Phase-plane-wshift}
\end{figure}

This section discuss the training algorithm for \DS and how to achieve a performance comparable to its full-precision counterpart. We propose to use sign-scale decomposition inspired by \cite{li2021s} which design for Shift networks with zero weights.
% , which design is to promote weight-sign switching during training. 
Our training method decomposes the discrete weights of DenseShift networks into two parts: a binary base term $\wsign$ and a PoT scale term which shifts the input activation for $S$ bits:
\begin{equation} \label{eq:wshift}
\wshift = \underbrace{\{2\mathbbm{1}(\wsign)-1\}}_{\mathclap{\text{Sign}}} \overbrace{2^{S_T}}^{\mathclap{\text{Scale}}},
\end{equation}
where $\mathbbm{1}(\cdot)$ is the Heaviside function mapping all positive values to one and the remaining to zero.

Next, we recursively re-parameterize the shifting parameter $S$ as a combination of $t$ binary variables  to address the weight freezing problem:
\begin{equation} 
    \label{eq:shift-scale}
    S_{0} = 0,  \   S_{t} = \mathbbm{1}(w_{t})  (S_{t-1} + 1),\quad\quad 1 \leq t \leq T. 
\end{equation}
In the following, we demonstrate the re-parameterization on positive $S$ values using $3$-bit case as an example, and the negative $S$ values can be obtained by adding a constant bias term.
We define a 3-bit DenseShift network with discrete weights $\wshift\in \{\pm1, \pm2, \pm4, \pm8\}$. This network can be re-parameterized as: 
% \begin{equation} 
%     \label{eq:shift-3-bit-S3}
%     S_{3} = \mathbbm{1}(w_{3})\{\mathbbm{1}(w_{2})\{\mathbbm{1}(w_1)+1\}+1\}.
% \end{equation}
% \begin{equation} \label{eq:shift-3-bit}
%     \wshift = \{2\mathbbm{1}(\wsign)-1\}2^{S_{3}},
% \end{equation}

\begin{eqnarray} 
S_{3} &=& \mathbbm{1}(w_{3})\{\mathbbm{1}(w_{2})\{\mathbbm{1}(w_1)+1\}+1\} \label{eq:shift-3-bit-S3},\\
\wshift &=& \{2\mathbbm{1}(\wsign)-1\}2^{S_{3}}. \label{eq:shift-3-bit}
\end{eqnarray}

Note that all the weights $\{\wsign, w_1, w_2, w_3\}$ are trained in full-precision. 
{By representing the original shift parameter $S_3$ with three full-precision parameters $w_1$, $w_2$, and $w_3$, we are projecting the optimization process from 1D space to higher-dimensional 3D space,  making the shift parameter easier to vary between different scales and thus easier to learn.}
{Compared to \cite{li2021s}, our approach eliminates the dense weight regularizer. This not only removes the need to tune an additional hyper-parameter but also simplifies the usage of our algorithm.}
While such training requires $(N+1)$ floating-point references, it is not as memory expensive as it appears, especially under 2/3-bit weight conditions. The memory is dominated by the activation with a large batch size during training.

% { To better explain why our weight re-parameterization approach performs better than the quantizer-based training method,} 
To better understand the advantages of our weight reparameterization approach over the quantizer-based training method,
in Fig. \ref{fig:Phase-plane-wshift}, we visualize the { optimization spaces of the shifting parameter $S$ defined i) by  a quantizer (figure left part) and ii) by a recursive product of binary variables (figure right part).}
In the quantizer's optimization space, the continuous weights accumulate at the three discontinuities of the quantizer as discussed in several earlier works \cite{nagel2022overcoming,liu2023oscillation,li2021s}. This observation implies that the weights attracted by the discontinuities could not move freely on the optimization space during training. 
In contrast, the weights in the optimization space defined by the recursive product of binary variables are gathered at the origin of the optimization space, and the value of the shifting parameter $S$ can vary easily according to the gradient update signal.
The visualization shows our re-parameterization promotes the shifting parameters $S$ to oscillate in an extensive range value during training instead of oscillating around the quantizer's threshold values. 
This design reduces the optimization space's rigidness and allowing the model to converge to a better solution \cite{li2021s}. 
% Note that using more weight parameters to represent the shift parameter $S$ almost doesn't affect the memory consumption, as the memory consumption is dominated by the activation calculations when training with a large batch size.

The local learning rate of individual parameter $\wsign$ in the proposed training scheme is significantly larger than the global learning rate on the discrete weight $\wshift$. 
Furthermore, we analyze the backward gradient computation of our proposed decomposition.
We estimate the backward gradient across the Heaviside function using a Straight-Through-Estimator (STE) \cite{bengio2013estimating}. The gradient update towards $\wsign$ is calculated as:
\begin{equation} 
    \label{eq:dLossdWsign}
    \frac{\partial \text{Loss}}{\partial \wsign}  = \frac{\partial \text{Loss}}{\partial \wshift} {2^{S_{T}}},
\end{equation}
where $2^{S_{T}} \in \{1,2,...,2^T\}.$ 
From Eq. \ref{eq:dLossdWsign} we observe that $2^{S_{T}}$ plays a role of learning rate scale factor, and it has an extensive value range. Hence, it may significantly impact the gradient update scale. 
Based on our observation, we propose a local learning rate re-scaling strategy to address this issue. We replace Eq. \ref{eq:dLossdWsign} with Eq. \ref{eq:dLossdWsign_rescale} during backward propagation to re-scale local gradient updates:
% \begin{equation} 
%     \label{eq:dLossdWsign_rescale}
%     \frac{\partial \text{Loss}}{\partial \wsign}  = \frac{\partial \text{Loss}}{\partial \wshift} {\sqrt{2^{S_{T}}}}.
% \end{equation}

\begin{equation} 
    \label{eq:dLossdWsign_rescale}
    \frac{\partial \text{Loss}}{\partial \wsign}  = \frac{\partial \text{Loss}}{\partial \wshift} {\sqrt{{S_{T}+1}}}.
\end{equation}

{ ImageNet experiments indicates that the local learning rate re-scaling  enhances the accuracy of 2bit and 3bit ResNet-18 models by 0.3\% and 0.7\%, respectively.}

% {\xinlin Our experiments verify the performance gain from this re-parameterization.}
% Our experiments demonstrate that this design significantly improves sub-4-bits \DS networks performance. {\xinlin This sentence refers to the ablation study}

% {\xinlin Optional text to explain the good performance of \Sthree reparameterization ?}

%However, the model is not yet transfer learnable at this stage. 

\subsection{Low-Variance Random Initialization for Transfer Learning}

\begin{figure}[t]
	\centering
    \includegraphics[trim=40 0 90 20,clip,width=0.49\linewidth]{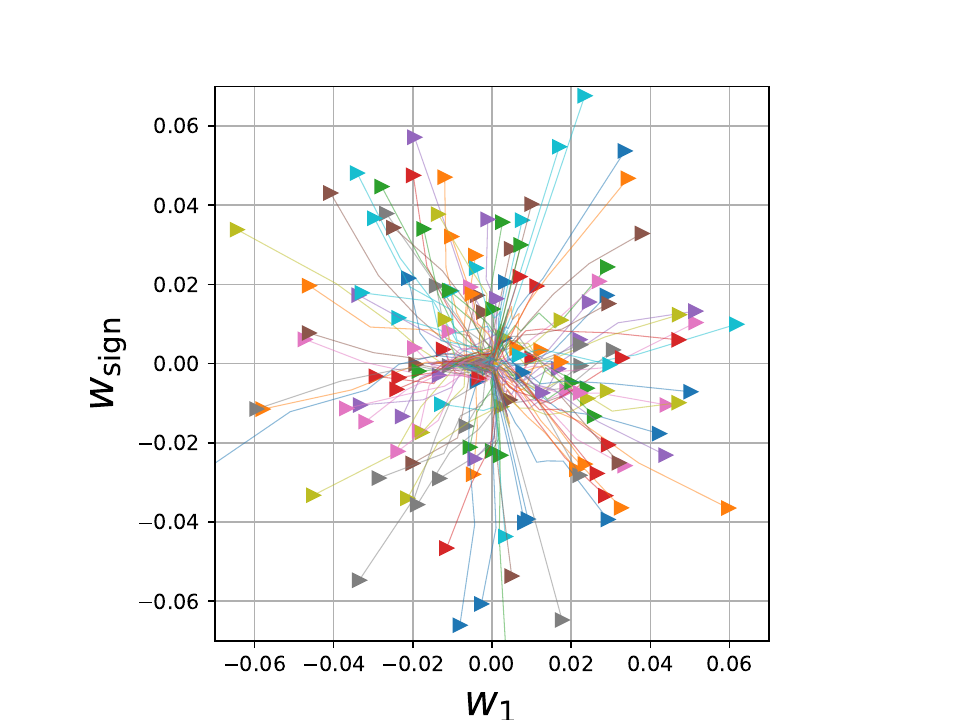}
    \includegraphics[trim=40 0 90 20,clip,width=0.49\linewidth]{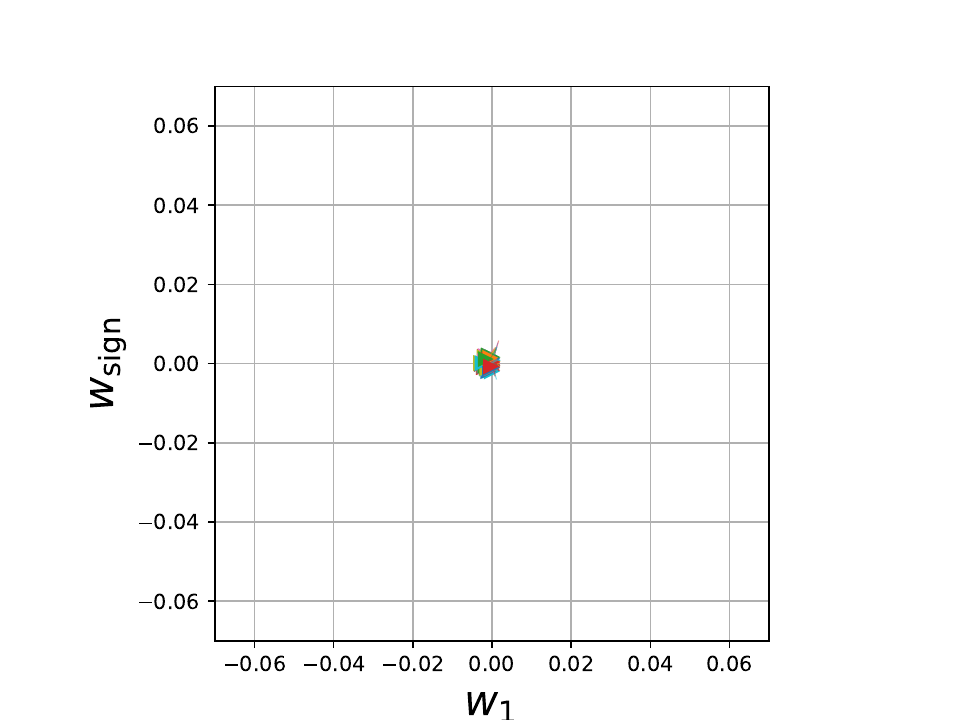}
	\caption{Re-parameterized weight trace visualization of a 2-bit DenseShift ResNet-18 trained on ImageNet dataset, the “triangle” indicates the initial point. These visualizations motivate us to develop low-variance initialization. Left panel: Kaiming initialization. Right panel: Low-variance initialization.} 
	 % \vspace{-4mm}
	\label{fig:weight_trace_shift2}
\end{figure}

%\subsubsection{Improving transfer learning performance with Low Variance Random Initialization}
We encountered difficulties when applying the above method in the transfer learning scenario. Most transfer learning tasks follow the following training paradigm: 
% {\rui match with above (i), (iii)}
i) Pre-train a backbone model on a large dataset;
ii) Remove and add new layers to the backbone model and randomly initialize the new layer;
iii) Finetune the new model on a downstream task.
During our experiments, we noticed that when we finetune a pre-trained DenseShift backbone with randomly-initialized DenseShift layers in an end-to-end manner, the model suffers from severe performance degradation or loss divergence. Such performance degradation also exists for existing Shift networks.

We analyzed the difference between the pre-trained weights and the randomly-initialized weights in the DenseShift model and noticed that the variance of the former is much lower than that of the latter. To better understand this phenomenon, we trained a 2-bit DenseShift network and noticed that when using the default Kaiming initialization \cite{he2015delving}, the weight values tend to gather towards the origin point of the re-parameterization space during the first few training epochs, as shown in figure \ref{fig:weight_trace_shift2}. 
This indicates that the initialized weight values are too far from the centre, an unwanted behaviour. More precisely, the weight values are far from the thresholds, meaning a greater initial gradient signal is needed to push them to pass the thresholds. 
In the transfer learning scenario, the backbone weight values are easier to change than the new Kaiming initialized layers.
% {\xinlin, leading to a different weight freezing issue in the transfer learning scenario.}
In fact, we argue that it is precisely this behavior that damages the pre-trained backbone during transfer learning. 
% Using a greater gradient signal to update the pre-trained backbone overshoots the existing weights and breaks them, thus reducing performance. This is because in the pre-trained backbone, the weights are grouped in the centre close to the threshold values, so a smaller gradient signal is required to fine-tune.

% The FCOS object detection experiments in Table \ref{tab:fcos} using quantized feature-pyramid networks are unstable and failed to converge during training without using our low-variance random initialization. 
To easily transfer a DenseShift network to other tasks, we suggest randomly initializing all re-parameterized variables with a small standard deviation. We name it low-variance random initialization.
Specifically, we chose a standard deviation of $10^{-3}$ for all the experiments in this paper. 
{  
The experiments in Sec. \ref{sec:transfer_learning} demonstrate that our low-variance random initialization strategy is required for achieving competitive performance on transfer learning tasks such as object detection and semantic segmentation. 
As evident from Table \ref{tab:ssd300}, the SSD 300 model shows a decrease in mAP without low-variance random initialization. 
Moreover, the FCOS object detection experiments, detailed in Table \ref{tab:fcos}, which employ quantized feature-pyramid networks, demonstrate instability and fail to converge during training without our proposed initialization.
}
In this paper, all \DS experiments use this initialization uniformly since we do not observed any negative impact on the model performance when training from scratch with our proposed initialization.

\begin{table}[t]
    \centering
    % \begin{tabular}{c|c|c|c|c}
    % \scalebox{0.85}
    %{
    \begin{tabular}{lllll}
    % \hline
    \toprule
        % \cline{1-5}
        \multicolumn{5}{c}{\textbf{ResNet-18}}\\ 
        Kernel & Methods  & W Bits & Init. & Top-1 (\%) \\ 
        \hline
        \multirow{2}{*}{Multi.} 
        & FP32 & 32 & R & 69.6 \\ 
        & TTQ \cite{zhu2016trained} & 2 & PT & 66.6 \\ 
        \hline
        \multirow{3}{*}{\shortstack[l]{Sum of \\ \\ Sign Flips}}
        & Lq-Nets \cite{zhang2018lq} & 2 & R &  68.0 \\ 
        & Lq-Nets \cite{zhang2018lq} & 3 & R &  69.3 \\ 
        & Lq-Nets \cite{zhang2018lq} & 4 & R &  70.0 \\ 
        \hline
        \multirow{9}{*}{Sign Flip}
        & BWN \cite{rastegari2016xnor} & 1 & R & 60.8\\
        & HWGQ \cite{cai2017deep} & 1 & R & 61.3 \\ 
        & BWHN \cite{hu2018hashing} & 1 & R & 64.3 \\ 
        & IR-net \cite{qin2020forward} & 1 & R & 66.5 \\
        & TWN \cite{li2016ternary} & 2 & R & 61.8 \\ 
        & LR-net \cite{shayer2017learning} & 2 & R & 63.5 \\ 
        & SQ-TWN \cite{dong2017learning} & 2 & R & 63.8 \\ 
        & INQ \cite{zhou2017incremental} & 2 & PT &  66.02 \\ 
        & \Sthree-Shift \cite{li2021s} & 2 & R & 66.37 \\ 
        \hline
        \multirow{11}{*}{\shortstack[l]{Shift + \\ Sign Flip}}
        & \bftab Ours & \bftab2 & LVR & \bftab 68.90\\ 
        & INQ \cite{zhou2017incremental} & 3 & PT &  68.08\\ 
        & INQ \cite{zhou2017incremental} & 4 & PT &  68.89\\ 
        & INQ \cite{zhou2017incremental} & 5 & PT &  68.98\\ 
        & DeepShift \cite{elhoushi2019deepshift} & 5 & R &  65.63\\ 
        & DeepShift \cite{elhoushi2019deepshift} & 4 & PT &  69.56\\ 
        & DeepShift \cite{elhoushi2019deepshift} & 5 & PT &  69.56\\ 
        & \Sthree -Shift \cite{li2021s} & 3 & R & 69.82\\ 
        & \bftab Ours & \bftab 3 & LVR & \bftab 70.62\\
        & STE \cite{przewlocka2022power} & 4 & PT &  69.98\\ 
        & \Sthree -Shift \cite{li2021s} & 4 & R & 70.47\\ 
        & \bftab Ours & \bftab 4 & LVR & \bftab 70.94\\
    \bottomrule
    \end{tabular}
    % }
    \caption{Comparison of SOTA methods using DenseShift ResNet-18 trained on ImageNet. Initialization defined as: R is Kaiming Normal Random initialization, PT is initialization with a Pre-Trained full-precision network, and LVR is our Low-Variance Random initialization. 
       }
     \vspace{-4mm}
    \label{tab:ResNet18-ImageNet}
\end{table}

\begin{table}[t]
    \centering
    % \begin{tabular}{c|c|c|c|c}
    % \scalebox{0.85}
    %{
    \begin{tabular}{lllll}
    % \hline
    \toprule
        \multicolumn{5}{c}{\textbf{ResNet-50}}\\ 
        % \cline{1-5}
        Kernel & Methods  & W Bits & Init. & Top-1 (\%) \\ 
        % \cline{1-5}
        % Kernel & Methods  & Bits & Init. & Top-1 (\%) \\ 
        \cline{1-5}
        Multi. & FP32 & 32 & R & 76.00 \\ 
        \cline{1-5}
        \multirow{2}{*}{\shortstack[l]{Sum of \\ Sign Flips}}
        & Lq-Nets \cite{zhang2018lq} & 2 & R &  75.10\\ 
        & Lq-Nets \cite{zhang2018lq} & 4 & R &  76.40\\ 
        \cline{1-5}
        \multirow{6}{*}{\shortstack[l]{Shift + \\ Sign Flip}} 
        & INQ \cite{zhou2017incremental} & 5 & PT &  74.81\\ 
        & DeepShift \cite{elhoushi2019deepshift} & 6 & PT &  75.29\\               
        & \Sthree -Shift \cite{li2021s} & 3 & R & 75.75\\ 
        & STE \cite{przewlocka2022power} & 4 & PT &  76.40\\ 
        & \bftab Ours & \bftab 2 & LVR & \bftab 75.62\\
        & \bftab Ours & \bftab 3 & LVR & \bftab 76.55\\ 
        & \bftab Ours & \bftab 4 & LVR & \bftab 76.53\\ 
        % & & & & \\
        % \cline{1-5}
        \toprule
        \multicolumn{5}{c}{\textbf{ResNet-101}}\\ 
        % \cline{1-5}
        Kernel & Methods  & W Bits & Init. & Top-1 (\%) \\ 
        \cline{1-5}
        Multi. & FP32 & 32 & R & 77.37 \\ 
        \cline{1-5}
        \multirow{3}{*}{\shortstack[l]{Shift + \\ Sign Flip}} 
        & \bftab Ours & \bftab 2 & LVR & \bftab 77.45\\
        & \bftab Ours & \bftab 3 & LVR & \bftab 77.93\\ 
        & \bftab Ours & \bftab 4 & LVR & \bftab 77.96\\ 
    \bottomrule
    \end{tabular}
    % }
    \caption{Comparison of SOTA methods using DenseShift ResNet-50/101 trained on ImageNet. 
    % Initialization defined as: R is Kaiming Normal Random, PT is Pre-Trained, LVR is Low-Variance Random. 
       }
     \vspace{-4mm}
    \label{tab:ResNet50-ImageNet}
\end{table}

\section{Experiments}

\subsection{ImageNet Classification}
\label{sec:ImageNet}
\textbf{Model and Dataset:} We benchmark our proposed method with different bit-widths. To verify the effectiveness and robustness, we apply DenseShift to ResNet-18/50/101 architectures and evaluate on ILSVRC2012 dataset \cite{deng2009imagenet} with data augmentation and pre-processing strategy proposed in \cite{he2016deep} and training strategy from \cite{li2021s}. Following \cite{rastegari2016xnor,liu2018bi,zhang2018lq}, all but the first convolution layers are quantized.\\
\textbf{Experiment Results:}
Results are shown in Table \ref{tab:ResNet18-ImageNet} and \ref{tab:ResNet50-ImageNet}. We compare our proposed method with SOTA low-bit multiplication-free networks using binary weights \cite{rastegari2016xnor,cai2017deep,hu2018hashing,qin2020forward}, ternary weights  \cite{li2016ternary,shayer2017learning,dong2017learning}, PoT weights \cite{zhou2017incremental,elhoushi2019deepshift,li2021s} and more complex kernel \cite{zhang2018lq}. We observe that DenseShift achieves SOTA performance on multiple network architectures and significantly outperforms the baseline with higher computational complexity.

\begin{table}[t]
    \centering
    % \scalebox{0.85}{
    \begin{tabular}{lllllll}
    \toprule
        \multirow{2}{*}{Methods} & \multicolumn{2}{c}{Quantized}  & \multicolumn{1}{c}{LVR} & \multirow{2}{*}{W Bits} &  \multirow{2}{*}{mAP} &  \\ 
        & Back & Head  & Init & & & \\ \hline
        FP32 & \d & \d & \d & 32 & 26.00\\ 
        \hline
        \multirow{3}{*}{\bftab Ours}
        & \cm & \d & \d & 3 &  26.23 \\ 
        & \cm & \cm & \cm & 3 &  25.75 \\
        & \cm & \cm & \d & 3 &   24.21 \\
    \bottomrule
    \end{tabular}
    % }
    \caption{DenseShift performs well on object detection. It confirms our low-variance initialization is necessary to keep high accuracy. We use DenseShift SSD300 v1.1 with ResNet-50 backbone finetuned on COCO object detection task. \emph{Back} is the backbone neural architecture, \emph{Head} is the detection head. Check-mark represents performing DenseShift quantization.}
     \vspace{-4mm}
    \label{tab:ssd300}
\end{table}

\begin{table}[t]
    \centering
    % \scalebox{0.85}{
    \begin{tabular}{lllllll}
    \toprule
        \multirow{2}{*}{Methods} & \multicolumn{3}{c}{Quantized}  & \multirow{2}{*}{W Bits}  & \multirow{2}{*}{mAP}\\ 
        & Back & FPN  & Head & & & \\ 
        \hline
        FP32 & \d & \d & \d & 32 & 39.0 \\ 
        \hline
        \multirow{9}{*}{\bftab Ours}
        & \cm & \d & \d & \multirow{3}{*}{2} &  39.3 \\
        & \cm & \cm & \d && 38.7 \\
        & \cm & \cm & \cm && 37.1 \\
        \cline{2-6}
        & \cm & \d & \d & \multirow{3}{*}{3} &  39.6 \\
        & \cm & \cm & \d && 39.3 \\
        & \cm & \cm & \cm && 37.7 \\
        \cline{2-6}
        & \cm & \d & \d & \multirow{3}{*}{4} &  39.8 \\
        & \cm & \cm & \d && 39.6 \\
        & \cm & \cm & \cm && 38.1 \\
    \bottomrule
    \end{tabular}
    % }
    \caption{DenseShift FCOS with ResNet-50 backbone finetuned on COCO object detection task.}
     \vspace{-4mm}
    \label{tab:fcos}
\end{table}
\subsection{Transfer Learning}
\label{sec:transfer_learning}
\textbf{Model and Dataset:} We use TorchVision implementation to verify the effectiveness and robustness of our proposed algorithm on transfer learning tasks. For object detection, we benchmark our proposed method on the bounding box detection track of MS COCO \cite{lin2014microsoft}. As proof of concept, we use SSD300 v1.1 \cite{liu2016ssd} with the obsolete VGG backbone replaced with ResNet-50 backbone. To demonstrate competitive performance, we use FCOS \cite{https://doi.org/10.48550/arxiv.1904.01355} with ResNet-50 backbone. For semantic segmentation, we benchmark our proposed method on a subset of MS COCO containing the 20 categories of Pascal VOC \cite{Everingham15}. We 
use DeepLab V3 \cite{chen2017rethinking} with ResNet-50 backbone architecture. The DenseShift ResNet-50 backbone is trained from the previous section and we compare against full-precision baselines using the same training strategy.\\
\textbf{Experiment Results:}
% We experiment with SSD300 \cite{liu2016ssd} v1.1, FCOS \cite{https://doi.org/10.48550/arxiv.1904.01355} and DeepLab V3 \cite{chen2017rethinking} with \DS ResNet-50 backbones for detection and segmentation tasks. We benchmark against TorchVision.
Tables \ref{tab:ssd300} and \ref{tab:fcos} illustrate that our 3-bit SSD300 and FCOS achieve similar performance to their full-precision counterparts. Table \ref{tab:deeplabv3} illustrate that our 3-bit DeepLab surpasses its full-precision counterpart. 
% Without using our low-variance random initialization, the FCOS experiments using quantized feature-pyramid networks are unstable and failed to converge during training. 

\begin{table}[t]
    \centering
    
    % \scalebox{0.85}{
    \begin{tabular}{lllllll}
    \toprule
        \multirow{2}{*}{Methods} & \multicolumn{2}{c}{Quantized}  & \multirow{2}{*}{W Bits}  & \multirow{2}{*}{mIoU} & Global \\ 
        & Back & Head  & & & Correct \\ 
        \hline
        FP32 & \d & \d & 32 & 66.4 & 92.4 \\ 
        \hline
        \multirow{6}{*}{\bftab Ours}
        & \cm & \d & \multirow{2}{*}{2} & 65.8 & 92.2\\
        & \cm & \cm & & 66.1 & 92.4\\
        \cline{2-6}
        & \cm & \d & \multirow{2}{*}{3} &  68.0 & 92.6\\
        & \cm & \cm & & 67.4 & 92.8\\
        \cline{2-6}
        & \cm & \d & \multirow{2}{*}{4} &  66.0 & 92.3\\
        & \cm & \cm & & 66.3 & 92.0\\
        % \cline{2-6}
    \bottomrule
    \end{tabular}
    % }
    \caption{DenseShift DeepLab V3 with ResNet-50 backbone finetuned on COCO semantic segmentation task.}
     \vspace{-4mm}
    \label{tab:deeplabv3}
\end{table}

\subsection{Speech Task}
\textbf{Model and Datasets:} To further demonstrate the generalization of DenseShift networks, we experiment on an end-to-end spoken language (E2E SLU) task with ResNet-18 architecture. We benchmark our method on the Fluent Speech Commands (FSC) dataset. The FSC dataset \cite{lugosch2019speech} comprised single-channel audio clips collected using crowd sourcing. Participants were requested to speak random phrases for each intent twice. The dataset contained 30,043 utterances spoken by 97 different speakers, each utterance contains three slots: action, object, and location. We considered a single intent as the combination of all the slots (action, object and location), resulting 31 intents in total.\\
\textbf{Experiment Results:} Results in Table \ref{tab:e2eslu} are benchmarked against full-precision and SOTA Shift networks performance. Our results demonstrates that our method can be applied to a field unrelated to the original CV field and can surpass full-precision performance as well.
% To further demonstrate the generalization of \DS networks, we experiment on an end-to-end spoken language (E2E SLU) task with ResNet-18. Results in table \ref{tab:e2eslu} are benchmarked against full-precision and SOTA performance. Our results demonstrates that our method can be applied to a field unrelated to the original CV field and can surpass full-precision performance as well.
% We use the same experiment setup as \cite{https://doi.org/10.48550/arxiv.2207.07497}. Table \ref{tab:e2eslu} proves that our \DS network can indeed be applied to a field unrelated to the original computer vision field.
\begin{table}[t]
    \centering
    
    % \scalebox{0.85}{
    \begin{tabular}{lcll}
    \toprule
        Method & W Bits & Val & Test \\
        \midrule
        \cite{https://doi.org/10.48550/arxiv.1904.03670} & 32 & 89.50 & 98.80\\
        \hline
        \multirow{2}{*}{\cite{https://doi.org/10.48550/arxiv.2207.07497}} & 2 & 90.66  & 98.41 \\ 
        & 3 & 90.31 & 98.41 \\
        \hline
        \multirow{2}{*}{\textbf{Ours}} & 2 & 90.73 & 98.60\\
        & 3 & 90.70 & 98.58\\
    \bottomrule
    \end{tabular}
    % }
    \caption{DenseShift ResNet-18 architecture on End-to-End Spoken Language Understanding}
     \vspace{-4mm}
    \label{tab:e2eslu}
\end{table}

\subsection{Quantized Activation}
Considering that Shift networks require fixed-point activation to achieve computational efficiency, we provide quantized activation experiments to verify the feasibility and find that 4-bit activation can maintain competitive performance on most CV tasks with PACT quantization \cite{choi2018pact}.
Results shown in Tables \ref{tab:ResNet-Pact}, \ref{tab:FCOS-Pact} and \ref{tab:DeepLab-Pact} demonstrate that DenseShift can attain similar performance to their full-precision counterparts. Hence, we believe DenseShift networks generalize well and are independent of other layers.
% To further verify the robustness of our proposed algorithm, we apply quantized activation and benchmark our previous CV experiments following \cite{https://doi.org/10.48550/arxiv.1805.06085}. 

\begin{table}[t]
    \centering
    % \begin{tabular}{c|c|c|c}
    % \scalebox{0.85}{
    \begin{tabular}{ccccc}
    \toprule
        Network & \multirow{2}{*}{A Bits} &
        \multicolumn{3}{c}{Top-1 Acc (\%)}\\
        Architecture & & 2 Bit  & 3 Bit  & 4 Bit \\ 
        % \hline
        % \multicolumn{5}{c}{\textbf{ResNet-18}}\\
        \hline
        \multirow{3}{*}{{ResNet-18}} & 32    & 68.90 & 70.62 & 70.94\\
        & 8     & 68.86 & 70.46 & 70.95\\
        & 4     & 68.56 & 70.00 & 70.35\\
        % \hline
        % \multicolumn{4}{c}{\textbf{ResNet-50}}\\
        \hline
        \multirow{2}{*}{{ResNet-50}} & 32    & 75.62 & 76.55 & 76.53\\
        & 4     & 75.27 & 76.18 & 76.63\\
        % \hline
        % \multicolumn{4}{c}{\textbf{ResNet-101}}\\
        % \hline
        % \multirow{2}{*}{{ResNet-101}} & 32    & 77.45 & 77.93 & 77.96\\
        % & 4     & 75.27 & 77.70 & 77.59\\
    \bottomrule
    \end{tabular}
     % }
     \caption{Quantized activation experiments using DenseShift ResNet architectures on ImageNet classification task. A-Bits    defined as activation bitwidth.}
      \vspace{-4mm}
    \label{tab:ResNet-Pact}
\end{table}

\begin{table}[t]
    \centering
    
    % \scalebox{0.85}{
    \begin{tabular}{cccclll}
    \toprule
        \multicolumn{3}{c}{Quantized} & 
        \multirow{2}{*}{A Bits} 
        % Activation
        & \multicolumn{2}{c}{mAP}\\ 
        Back & FPN & Head  &  & 3 Bit & 4 Bit\\ 
        \midrule
        \cm & \cm & \d  & 32 & 39.3 & 39.6\\ 
        \cm & \cm & \d  & 4  & 39.6 & 39.6\\
        \cm & \cm & \cm & 32 & 37.7 & 38.1\\ 
        \cm & \cm & \cm & 4  & 37.8 & 37.8\\
    \bottomrule
    \end{tabular}
    % }
    \caption{Quantized activation experiments using DenseShift FCOS with ResNet-50 backbone finetuned on COCO object detection task.}
     \vspace{-4mm}
    \label{tab:FCOS-Pact}
\end{table}

\begin{table}[t]
    \centering
    
    % \scalebox{0.85}{
    \begin{tabular}{cccllll}
    \toprule
        \multicolumn{2}{c}{Quantized}  &
        \multirow{2}{*}{A Bits} 
        % Activation
        &  \multicolumn{2}{c}{mIoU}\\ 
        Back & Head  & & 2 Bit & 3 Bit \\ 
        \hline
        \cm & \d  & 32 & 65.8 & 68.0\\ 
        \cm & \d  & 4  & 64.9 & 66.3\\
        \cm & \cm & 32 & 65.5 & 66.4\\ 
        \cm & \cm & 4  & 65.2 & 65.9\\
    \bottomrule
    \end{tabular}
    % }
    \caption{Quantized activation experiments using DenseShift DeepLab V3 with ResNet-50 backbone finetuned on COCO semantic segmentation task.}
    \label{tab:DeepLab-Pact}
\end{table}

{ 
\subsection{Ablation Study}

{\bftab Training epochs.} As highlighted in prior studies  \cite{alizadeh2018empirical,courbariaux2015binaryconnect,tang2017train,li2021s,xu2023resilient}, the training of binary variables necessitates additional epochs due to the instability arising from frequent sign variations.  Our experiments in Table \ref{tab:ResNet18-Epochs} verified the impact of training epochs on the network performance for the re-parameterized training of \emph{DenseShift} networks. Apart from the number of epochs, all other model settings and training strategies adhere to Sec. \ref{sec:ImageNet}.
}

\begin{table}[t]
    \centering
    \begin{tabular}{ccccc}
    \toprule
        Network & \multirow{2}{*}{W Bits} &
        \multicolumn{3}{c}{Training Epochs}\\
        Architecture & & 90 & 150  & 200 \\ 
        \hline
        \multirow{2}{*}{{ResNet-18}} & 2    & 67.36 & 68.41 & 68.90\\
        & 3     & 69.30 & 69.91 & 70.62\\
    \bottomrule
    \end{tabular}
     \caption{Ablation study on training epochs using DenseShift ResNet-18 architectures on ImageNet classification task.}
      \vspace{-4mm}
    \label{tab:ResNet18-Epochs}
\end{table}

\section{Conclusion}
We present \DS with zero-free shifting and sign-scale decomposition for constructing high-performance low-bit Shift networks with high training and inference efficiency. For the first time, Shift networks support inference with non-quantized floating-point activations and achieve performance gain on general hardware such as ARM CPU.
Furthermore, we propose a low-variance random initialization strategy that enhances the performance of DenseShift networks in transfer learning, allowing the networks to adapt to various tasks without significant performance degradation.
Our extensive experiments on various tasks demonstrate that DenseShift networks outperform current state-of-the-art Shift networks in classification tasks and achieve comparable performance to full-precision models in object detection and semantic segmentation tasks. This breakthrough represents a significant advancement for low-bit Shift networks.

{
\section*{Acknowledgement}
We thank the continuous support of Boxing Chen and Wulong Liu throughout this project. We also appreciate Masoud Asgharian's help in revising the proof of the theorems.
}

{\small
\bibliographystyle{ieee_fullname}
\bibliography{egbib}
}

~\newpage
~\newpage
\appendix
% \section{Supplementary Material}

% The supplementary material contains three sections. 
% \textbf{Section \ref{sec:theorem}} provides two theorem proofs of DenseShift networks. Theorem 1 shows every shift network has a dense shift representation. Theorem 2 shows dense shift networks are universal approximators.
% \textbf{Session \ref{sec:correction}} provides error correction to our submitted paper draft.
% \textbf{Session \ref{sec:code}} describes our demo code for the reproducibility check.

% \subsection{Proof of Theorems} 
\section{Proof of Theorems}
\label{sec:theorem}
\begin{theo}
Suppose a shift neural network in the form $\hat f_\w(\x)=\sum_{j=1}^J\sum_{k=1}^{d_j} w^j_{k} h^j_k(\x)
 ,$ where $w^j_k \in \{0\} \cup \{\pm 2^p\}$ approximates the true unknown function $f(\x), \x\in \real^d.$ There is another dense  neural network with $w'^j_k \in \{\pm 2^p\}$ that has a similar form $\hat f_{\w'} (\x)=\sum_{j=1}^J\sum_{k=1}^{d'_j} \w'^j_{k} h'^j_k(\x)$, such that  $\forall \x\in\real^d, \hat f_\w(\x) =\hat f_{\w'} (\x) .$
\end{theo}
\begin{proof}
The proof is straightforward by isolating zero shifts, and recreating these null weights in a larger dense shift network with opposite weight signs.
Define the $j^{\text{th}}$ layer as
$$h^j(\x) = a(\W^j h^{j-1}(\x)+\b^j), $$
where $h^0(\x) = \x$, $J$ is the total number of layers each of output dimension $d_j$ and $\W$ of size $d_j\times d_{j-1}$ can be  a Toeplitz matrix for a convolutional layer, $a(\cdot)$ is the activation function, and $\b$ is the bias term. Define the shift network approximation of $f(\x)$ as
\begin{eqnarray*}
\hat f_\w(\x)=\sum_{j=1}^J\sum_{k=1}^{d_j} w^j_{k} h^j_k(\x)
\end{eqnarray*}
in which $w^j_k \in \{0\} \cup \{\pm 2^p\}$  defines a shift network. We re-create an equivalent desne shift network by isolating null weights $w^j_k = 0$ and replacing them with a larger dense shift network of an arbitrary weights but with opposite signs. Now suppose $\w_0=\{k\mid w^j_{k}=0\}$ and $\w_1 =\{k\mid w^j_k \neq 0\}$ where $\w=[\w_0^\top \w_1^\top]$, $d_j=d_{0j}+d_{1j}$
\begin{eqnarray*}
\hat f_\w(\x)=\sum_{j=1}^J\left[\sum_{k=1}^{d_{0j}} w^j_{0k} h^j_{0k}(\x) + \sum_{k=1}^{d_{1j}} w^j_{1k} h^j_{1k}(\x)\right].
\end{eqnarray*}
Assume $\tilde \w\in \{\pm 2^p\}$ is a nonzero shift arbitrary vector of elements $\tilde w_{0k}$, 
\begin{eqnarray*}
&&\sum_{j=1}^J\left[\sum_{k=1}^{d_{0j}} \w^j_{0k} h^j_{0k}(\x) + \sum_{k=1}^{d_{1j}} \w_{1k} h^j_{1k}(\x)\right]\\
&=& \sum_{j=1}^J\left[\sum_{k=1}^{d_{0j}} \tilde \w^j_{k} h^j_{0k}(\x) - \tilde \w^j_{k} h^j_{0k}(\x) + \sum_{k=1}^{d_{1j}} \w_{1k} h^j_{1k}(\x)\right]. \\
\end{eqnarray*}
By defining $\w'=[\tilde \w,  -\tilde \w,  \w_0]$ of increased size $d'_j={d_{0j}+d_j\geq d_j }$, one may rearrange terms and rewrite the neural approximate as
\begin{eqnarray*}
\hat f_{\w'} (\x)=\sum_{j=1}^J\sum_{k=1}^{d'_j} \w'^j_{k} h'^j_k(\x), 
\end{eqnarray*}
where $h'^j_k$ is either $h^j_{0k}$ or $h^j_{1k}$ depending on the dense shift weight $\w'^j_k \in \{\pm 2^p\}$.
\end{proof}
\begin{theo}
Dense shift network with a Lischitz activation function is a universal approximator on a compact set $K$ for any measurable continuous function $f\in C(K)$ with respect to the measure $\mu$, given that its weight and activations $\w,h(\x)$ remain close to the regular network in the following sense   
$$\int_K  \left(\sum_{j=1}^J\sum_{k=1}^{d_j} \left(\w^j_{k}h^j_k(\x) -  \w'^j_{k}h'^j_k(\x)\right)\right) ^2d\mu <{\epsilon \over 4},$$
where  $\epsilon\over 4$ is the approximation quality of the regular neural network,  $\left(\w, h^j(\x)\right)$ and $\left(\w', h'^j(\x)\right)$ are the weight and activations of the regular and DenseShift networks in layer $j$, respectively.
\end{theo}
\begin{proof}
It is well-known that shallow networks are universal approximator  \cite{Hornik_UniversalMLP_1991} as well as deep networks \cite{Zhou_UniversalDeepCNN_2020}. These  results holds in infinity norm, so is also valid in $\ell_p$ norm with $p<\infty$. For the simplicity of the mathematical mechanics here we only focus on  the multilayer perceptron \cite{Hornik_UniversalMLP_1991} on $\ell_2$ norm
\begin{eqnarray}
\norm{f - \hat{f}} =\int_K \left | f(\x) - \hat{f}(\x) \right|^2  d\mu \label{eq:approxorigin}
\end{eqnarray}
Suppose $\hat f_\w$ is a  real weight neural network and  $\hat f_{\w'}$ is a dense shift version of the same network $\hat f_\w$, of course with weights $\w'\in \{\pm 2^p\}$.
{
\begin{eqnarray}
&& \norm{{f}- \hat f_{\w'}} \nonumber\\
&=& \int_K \left| f(\x) - \hat{f}_{\w'}(\x)\pm\hat{f}_\w(\x)  \right|^2  d\mu \\
&=& \int_K  |f(\x) -\hat f_\w(\x)  |^2 d\mu \label{eq:firstterm}\\
&&+ \int_K  |\hat f_\w(\x) - \hat f_{\w'} (\x)|^2d\mu \label{eq:firstsecondterm} \\
&&+2\int_K  |[ f (\x)- \hat f_\w(\x)  ] [\hat f_\w(\x) - \hat f_{\w'}(\x)]| d\mu. \label{eq:thirdterm}
\end{eqnarray}
}
 In order to show that shift networks are universal approximator it is enough to show that $\norm { f-\hat f_{\w'}}$ is bounded by an arbitrarily small $\epsilon>0$.
 The first term \eqref{eq:firstterm} is bounded by $\epsilon \over 4$   \cite{Hornik_UniversalMLP_1991}. The second term \eqref{eq:firstsecondterm} is bounded
 given the shift net closeness assumption 
\begin{eqnarray*}
&& \int_K  |\hat f_\w(\x) - \hat f_{\w'} (\x)|^2d\mu \\
&=& 
\int_K  \left(\sum_{j=1}^J\sum_{k=1}^{d_j} \w^j_{k}h^j_k(\x) - \sum_{j=1}^J\sum_{k=1}^{d_j} \w'^j_{k}h'^j_k(\x)\right) ^2d\mu \\
&<& {\epsilon\over 4}
%&\leq&  \int_K  \left(\sum_{j=1}^J\sum_{k=1}^{d_j} \left(L\w^j_{k}\x^j_k -  L\w'^j_{k}\x'^j_k\right)\right) ^2d\mu \\
% &\leq&  \int_K  \left(L \sum_{j=1}^J\sum_{k=1}^{d_j} \left(\w^j_{k}\x^j_k -  \w'^j_{k}\x'^j_k\right)\right) ^2d\mu < {\epsilon\over 4},
\end{eqnarray*}
The last term \eqref{eq:thirdterm} is bounded by ${\epsilon \over 2}$ thanks to the Cauchy-Schwartz inequality. So \eqref{eq:approxorigin} is bounded by $\epsilon$ by merging the pieces together.
\end{proof}

% \subsection{Corrections} \label{sec:correction}
% % \textbf{Line 150: } They remain computational costly as \textbf{more operations} are used per kernel. \\
% % \textbf{Table 1: } 2bit Dense Shift networks use \textbf{Shift + Sign Flip} kernel for inference.
% \textbf{Equation \ref{eq:float_val}:} 
% \begin{equation} 
%     \label{eq:float_val_cor}
%     % {\textbf{Val}}_{float} 
%     \Val(x) = (-1)^{x_{\textbf{s}}} \times 2^{x_{\textbf{e}} + \textbf{e}_{\textbf{bias}}} \times (1 + \frac{x_{\textbf{m}}}{2^{\textbf{m}_{\textbf{bits}}} - 1}),  
% \end{equation}

% \subsection{Demo code \& Open-source plan} \label{sec:code}

% The supplementary material contains a 3-bit DenseShift training demo code using ResNet-18 on the ImageNet classification task for reproducibility check. However, the pre-trained checkpoints are not included due to the size limitation on the supplementary materials. And the ICCV double-blind policy discouraged authors from providing links to websites created for submission to offer code or data. We provide an output log example of the training demo code as compensation. More details in \emph{README.md}.\\
% We plan to open-source our project. Due to our internal open-source review process, we can not promise the open-source code will be available before the camera-ready deadline.

\end{document}